\documentclass[letterpaper, 10 pt, conference]{IEEEconf}
\usepackage{hyperref}
\usepackage[utf8]{inputenc}
\usepackage[T1]{fontenc}
\usepackage{graphicx}
\usepackage{longtable}
\usepackage{wrapfig}
\usepackage{rotating}
\usepackage[normalem]{ulem}
\usepackage{amsmath}
\usepackage{amssymb}
\usepackage{capt-of}
\usepackage{hyperref}
\hypersetup{colorlinks=true,allcolors=blue}
\usepackage{cite}
\IEEEoverridecommandlockouts % This command is only needed if you want to use the \thanks command
\usepackage{acronym}
\acrodef{QOI}{quantity of interest}
\acrodef{AS}{adaptive sampling}
\acrodef{GPR}{Gaussian Process Regression}
\acrodef{GP}{Gaussian Process}
\acrodefplural{GP}{Gaussian Processes}
\acrodef{MTGP}{Multi-Task Gaussian Process}
\acrodefplural{MTGP}{Multi-Task Gaussian Processes}
\acrodef{STGP}{Single-Task Gaussian Process}
\acrodefplural{STGP}{Single-Task Gaussian Processes}
\author{Nicholas Harrison, Nathan Wallace, and Salah Sukkarieh \thanks{All authors are with the Australian Centre for Field Robotics, University of Sydney, NSW, AU. nicholas.harrison@sydney.edu.au, nathan.wallace@sydney.edu.au, salah.sukkarieh@sydney.edu.au}}
\date{\today}
\title{\LARGE \bf Automated Testing of Spatially-Dependent Environmental Hypotheses through Active Transfer Learning}
\hypersetup{
 pdfauthor={Nicholas Harrison, Nathan Wallace, and Salah Sukkarieh \thanks{All authors are with the Australian Centre for Field Robotics, University of Sydney, NSW, AU. nicholas.harrison@sydney.edu.au, nathan.wallace@sydney.edu.au, salah.sukkarieh@sydney.edu.au}},
 pdftitle={\LARGE \bf Automated Testing of Spatially-Dependent Environmental Hypotheses through Active Transfer Learning},
 pdfkeywords={},
 pdfsubject={},
 pdfcreator={Emacs 29.2 (Org mode 9.7)}, 
 pdflang={English}}
\begin{document}

\maketitle
\bstctlcite{IEEEexample:BSTcontrol}

\begin{abstract}
The efficient collection of samples is an important factor in outdoor information gathering applications on account of high sampling costs such as time, energy, and potential destruction to the environment. Utilization of available a-priori data can be a powerful tool for increasing efficiency. However, the relationships of this data with the quantity of interest are often not known ahead of time, limiting the ability to leverage this knowledge for improved planning efficiency. To this end, this work combines transfer learning and active learning through a Multi-Task Gaussian Process and an information-based objective function. Through this combination it can explore the space of hypothetical inter-quantity relationships and evaluate these hypotheses in real-time, allowing this new knowledge to be immediately exploited for future plans. The performance of the proposed method is evaluated against synthetic data and is shown to evaluate multiple hypotheses correctly. Its effectiveness is also demonstrated on real datasets. The technique is able to identify and leverage hypotheses which show a medium or strong correlation to reduce prediction error by a factor of 1.4--3.4 within the first 7 samples, and poor hypotheses are quickly identified and rejected eventually having no adverse effect.
\end{abstract}

\section{Introduction}
\label{sec:org3122a32}

Many motivations exist for gathering environmental data across various outdoor applications. These may be driven by 1) scientific objectives: a desire to learn and characterize an unknown environment or 2) operational concerns: leveraging this new knowledge to refine plans and increase efficiency. This process not only underpins effective planning and decision-making, but it also facilitates the minimization of resource expenditure, be it time, energy, or environmental impact.

\begin{figure}[t]
\centering
\includegraphics[width=0.8\linewidth]{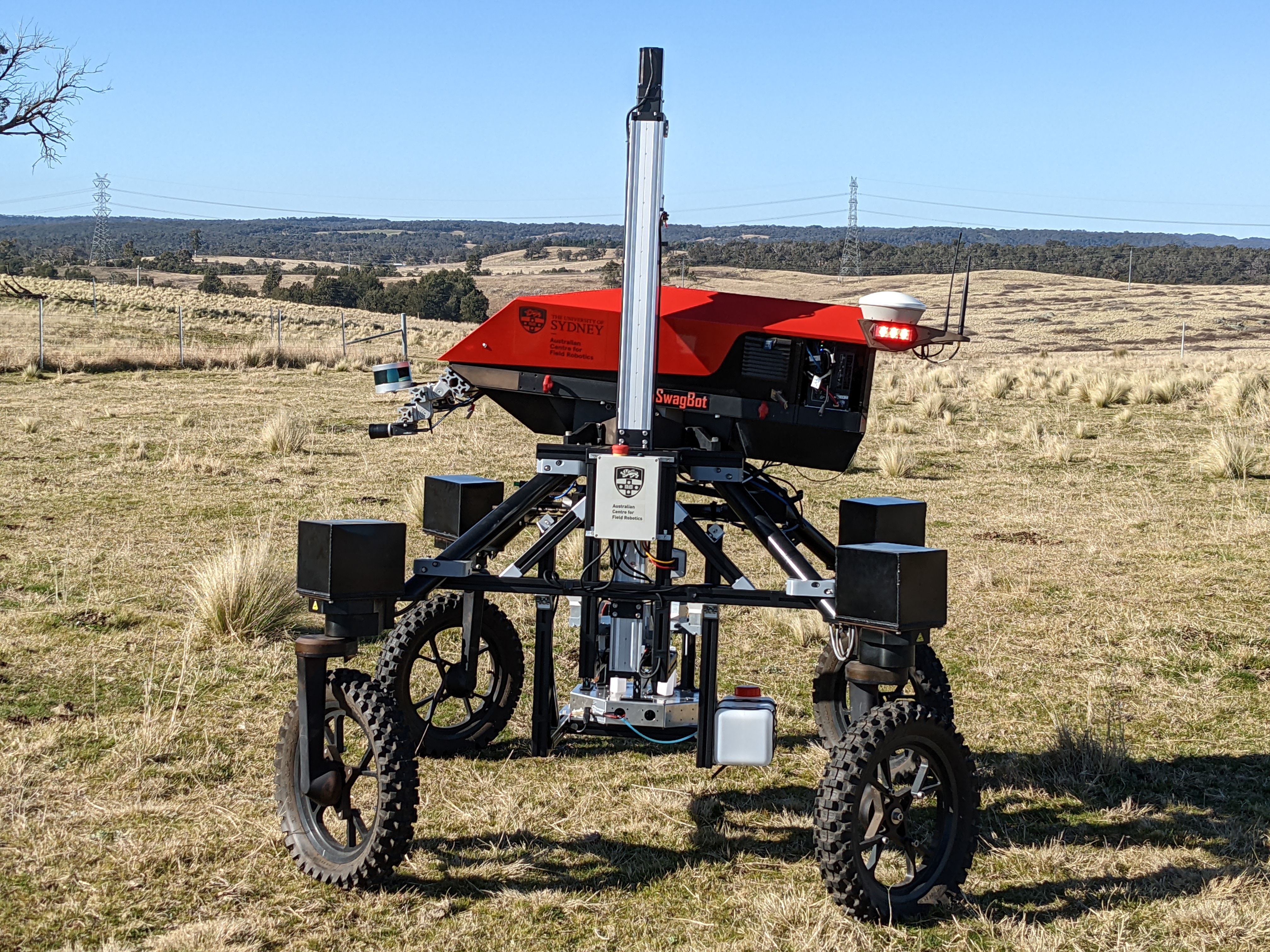}
\caption{\label{fig:swagbot}The ACFR's Swagbot autonomously collecting and analyzing soil samples.}
\end{figure}

A prime example of such strategies being applied in an agricultural context is our recent work on using an autonomous rover to collect soil samples and test them onboard for Nitrogen, Phosphorous, and Potassium content (Fig. \ref{fig:swagbot}). Understanding these values is important for farmers to more efficiently manage their fields and increase yields \cite{kwon19effec_nitr}. The process to extract the soil and then pass it through the stages of an onboard lab is complex and time-intensive. This results in only sparse data available and limited capacity to collect new data. In scenarios like these, it becomes a priority to gain the most information from the fewest samples as possible.

\begin{figure}[t]
\centering
\includegraphics[width=1.0\linewidth]{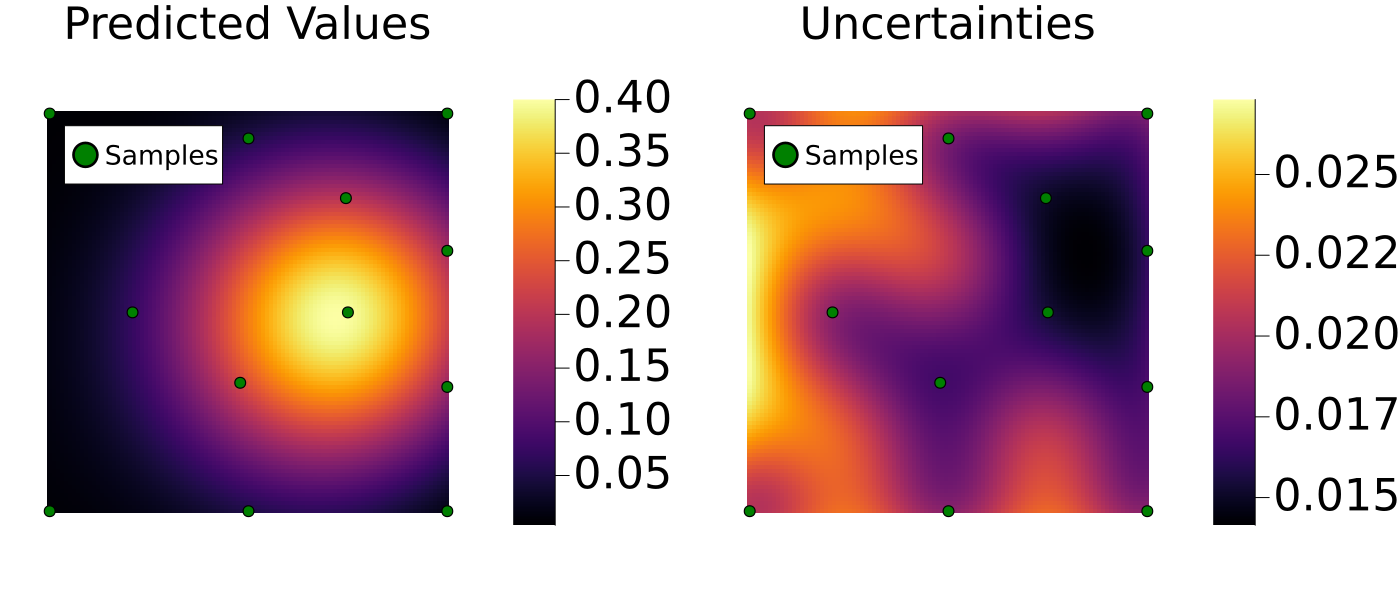}
\caption{\label{fig:example_gp}Example \ac{GP} maps. Yellow and black indicate high and low values, respectively, and green dots show sample locations. \acp{GP} produce smooth interpolations between measurements and higher uncertainty away from sample locations.}
\end{figure}

Incorporating prior information helps to address this problem. For any given \ac{QOI}, for example soil Nitrogen content, other sources of information may exist in the environment about its distribution. Environmental quantities are often spatially-dependent, meaning their measured values depend on where they are in space. A useful technique to exploit this is the \ac{GP}, known as kriging in geostatistics. This nonparametric machine learning technique uses a covariance function to define how relations in space dictate relations in measurement values \cite{rasm06gaus_proc}. It can give an estimate for a quantity at any point in space conditioned on sparse samples. See example maps produced from a \ac{GP} in Fig. \ref{fig:example_gp}.

Additionally, correlations often exist between the values of different quantities. One example is the relationship between Nitrogen, soil moisture, and temperature \cite{gunt12effec_mois}. These relationships can be exploited through the technique of transfer learning. Though transfer learning is often applied to image and text classification \cite{zhuan21comp_surv,Weiss_2016}, the concept is also useful in domains such as this one. In the context of this work, transfer learning improves predictions of a \ac{QOI} by using learned patterns in related quantities. This can also be realized through \acp{GP} through a ``multi-task'' covariance function that accommodates data from multiple sources.

To improve the efficiency of learning these relationships, the technique of active learning can be applied, which involves the process of choosing better samples to take based on their potential informativeness. \acp{GP} give values of uncertainty about the estimates they produce, which can be used to calculate which samples will yield the most information about the \ac{QOI}.

Beyond characterization efficiency, combining these two learning techniques has a valuable scientific extension: learning the true global relationships between quantities in the environment. Relationships can often be guessed (i.e. hypothesized) intuitively by a human, but they may or may not be valid for the current region. In this work, hypotheses are defined as a linear dependence between a \ac{QOI} and another quantity. Transfer learning can represent such hypotheses, and the combination with active learning can test them and learn more accurate ones. Accurately-learned relationships can be confidently re-used in future applications.

One additional and significant aspect of dealing with sparse data from multiple quantities is that samples from the different quantities may not line up in space, which is known as non-collocated data. This aspect makes calculating correlations directly between quantities impossible. An intelligent use of \acp{GP} can handle this too by exploiting their property to interpolate values continuously through space. In effect, a spatially-dependent correlation is calculated.

\subsection{Problem Definition and Contributions}
\label{sec:org08cd937}
In this work, we consider the task of autonomously characterizing a single \ac{QOI} within a constrained area, such as a section of rangeland. The ground truth data for this quantity are not known by the robot, but it does have sparse data of other quantities in the environment that are hypothesized to be linearly correlated with the \ac{QOI}. The robot can travel within the region, take samples of the \ac{QOI}, and calculate predictions of the \ac{QOI}'s distribution as well as the quality of the hypotheses. The robot must sequentially select which sample locations will yield the most information about the \ac{QOI} and proposed hypotheses.

The main contribution of this work is a method which
\begin{itemize}
\item calculates the quality of multiple environmental hypotheses simultaneously and in real-time from sparse, non-collocated data
\item learns from good hypotheses, improving prediction accuracy
\item disregards poor hypotheses, incurring no loss
\end{itemize}

After a review of related research, this article will explain the methods for the combined active transfer learning.
\section{Related Work}
\label{sec:org56c79d5}

Within the domains of information gathering in natural environments and automated hypothesis testing, a few works are especially relevant.

Furlong investigated a form of hypothesis testing, picking new samples based on greater entropy in the hypothesis space \cite{furl18fora_pros}. Their algorithm identified which out of a number of hypotheses was the most true for predicting a spatial map of binary variables. Since they were binary rather than continuous, each hypothesis was either completely true or completely false at each location. They assumed locations in space were independent and compared hypotheses point-by-point, adding a spatial reward to spread out samples.

Also in the area of hypothesis testing, Burger \cite{burg20mobil_robot} developed a robotic chemist: a system to find an optimal chemical solution using a batched Bayesian search algorithm. The algorithm used \ac{GP} mean and variance to pick potentially favorable combinations of compounds to test. It indirectly tested hypotheses related to the compounds it was given, but it didn't produce a measure of the validity of each hypothesis. This is in part because it didn't include a model for the relationship between the resulting compound and its constituent parts.

For information gathering, a number of works use information-theoretic measures to guide a robot to locations of greatest informativeness. A few examples of these include information gain \cite{aror18onlin_mult}, upper confidence bound \cite{marc12bayes_intel}, ergodic theory \cite{edel20ergod_traj}, and upper-bounded differential entropy \cite{cand20plan_rover}. \acp{GP} are common in these scenarios because they estimate both values and uncertainties --- useful for quantifying current and potential information and defining objective functions \cite{fuhg20stat_of,shah16takin_human}.

In addition, many works use various forms of multi-quantity \acp{GP} to share information and improve predictions of natural phenomena. For example, \cite{edel20ergod_traj} and \cite{cand20plan_rover} incorporated remote spectroscopic data as part of the input space in the \ac{GP} covariance function to enhance local predictions. In contrast, \cite{leen90empl_elev} and \cite{meht21adap_samp} use forms that predict multiple related output values from a common input space. In all of these, the structure chosen allowed correlations, but they didn't provide a way to globally evaluate the relationship between the quantities, whether strong or weak.

Overall, a method is needed to autonomously measure the spatial dependence of two quantities in a natural environment and adapt learning to that measure.
\section{Methods}
\label{sec:orgec15856}

\subsection{Model Learning}
\label{sec:org260524b}

Predictions and uncertainties across the search region are modeled using a \ac{GP}, which is defined by its mean and covariance functions. For the base of the covariance function, we use the squared exponential. Its simplicity leads to easier derivation and lower computational cost, and it is applicable to spatially-continuous phenomena found in nature. It takes the form
\begin{equation} \label{eq:basefunc}
cov(\mathbf{x}_i, \mathbf{x}_j | \sigma_f^2, \sigma_l^2) = \sigma_f^2 \exp \left(- \frac{|| \mathbf{x}_i - \mathbf{x}_j ||^2}{2 \sigma_l^2} \right) ,
\end{equation}
where \(\mathbf{x}_i\) and \(\mathbf{x}_j\) are sample locations, and \(\sigma_f\) and \(\sigma_l\) act as the signal standard deviation and characteristic length-scale, respectively. This is a common single-quantity covariance function. It causes locations closer in space to have greater covariance, which means their sample values are assumed to be closer.

To transfer information between quantities, something must additionally represent the amount of covariance between them. We follow the pattern of Bonilla in replacing the signal variance with a free-form covariance matrix \cite{boni08mult_task},
\begin{equation} \label{eq:covfunc}
cov(\mathbf{x}_{i,u}, \mathbf{x}_{j,v} | A, \sigma_l^2) = A_{uv} \exp \left(- \frac{|| \mathbf{x}_{i,u} - \mathbf{x}_{j,v} ||^2}{2 \sigma_l^2} \right) ,
\end{equation}
where \(u\) and \(v\) are quantity indices, and the matrix element \(A_{uv}\) is the covariance between two quantities. The covariance function's two factors each serve a role. The exponential factor serves as a spatial correlation between samples ranging from \(0\) to \(1\). The matrix \(A\) acts as a covariance between entire quantities, i.e. a ``quantity-covariance''.

\begin{figure}[t]
\centering
\includegraphics[width=0.65\linewidth]{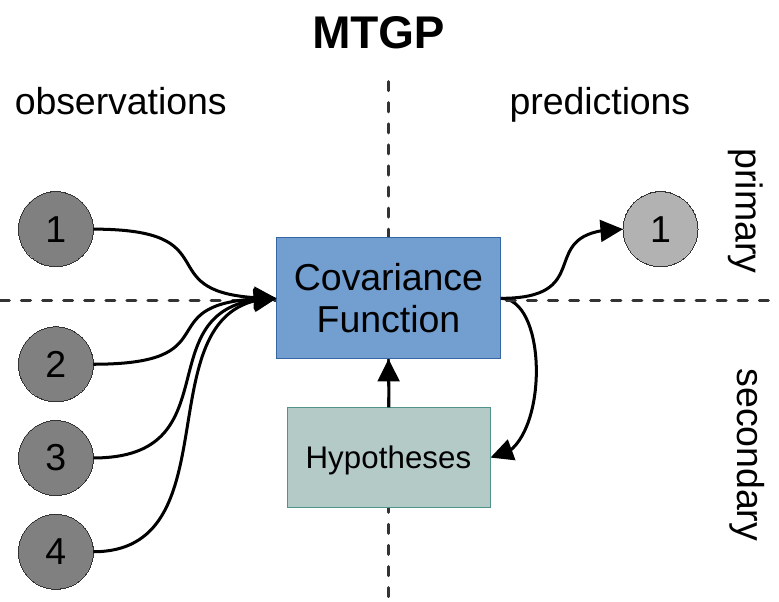}
\caption{\label{fig:mtgp_diagram}Abstract depiction of the \ac{MTGP} architecture. Gray circles represent values for each of the quantities, with \(1\) labeling the \ac{QOI}. The covariance function allows transfering information between quantities and representing how much they correlate (hypotheses).}
\end{figure}

For the mean function, we define a quantity-wise constant mean. That is, for each quantity, it will return the average of all the sample values of that quantity:
\begin{equation} \label{eq:meanfunc}
m(\mathbf{x}_{i,u}) = \mathbb{E} \left[ y_{.,u} \right]
\end{equation}
This piece, along with the covariance, is essential to producing hypothesis evaluations as described in the next section. See Fig. \ref{fig:mtgp_diagram} for an abstract idea of the \ac{MTGP} architecture.

To produce a valid positive semidefinite covariance matrix for \(A\), transpose multiplication with a lower-triangular matrix is used (example with 4 quantities):
\begin{align} \label{eq:covmat}
\begin{split}
A &= L L^T \\
L &=
\left[ \begin{array}{cccc}
a_1 & 0   & 0   & 0 \\
a_2 & a_3 & 0   & 0 \\
a_4 & a_5   & a_6 & 0 \\
a_7 & a_8   & a_9   & a_{10} \\
\end{array} \right],
\end{split}
\end{align}

These hyperparameters \(a_i\) can be chosen arbitrarily to cause or prevent correlations between the quantity predictions. All hyperparameter values are trained by maximizing the log marginal likelihood of the data given the hyperparameters as described in \cite{rasm06gaus_proc},
\begin{align} \label{eq:likelihood}
\begin{split}
\log p(\mathbf{y}|X,\theta) &= - \frac{1}{2} \left[ \tilde{\mathbf{y}}^T \Sigma^{-1} \tilde{\mathbf{y}} + \log|\Sigma| + n \log 2\pi \right] \\
\Sigma &= K(X, X | \theta ) + \sigma_n^2 I \\
\tilde{\mathbf{y}} &= \mathbf{y} - m(X) ,
\end{split}
\end{align}
where \(K(X,X|\theta)\) is the covariance matrix between all training data produced using (\ref{eq:covfunc}), \(m(X)\) is the array of means produced using (\ref{eq:meanfunc}), and \(\sigma_n^2\) is the hyperparameter for the noise variance and is also trained.

\subsection{Hypothesis Evaluation}
\label{sec:orge72bfc8}

Each hypothesis tested in this work is defined as the claim that a prior quantity is linearly dependent with the \ac{QOI}. To evaluate the validity of a hypothesis, we compute a coefficient of determination using the quantity-covariance matrix part of the covariance function (\ref{eq:covmat}). The correlation coefficient between two quantities can be calculated as
\begin{equation} \label{eq:}
r_{uv} = \frac{\sigma_{uv}}{\sigma_{u} \sigma_{v}} = \frac{A_{uv}}{\sqrt{A_{uu}}\sqrt{A_{vv}}} ,
\end{equation}
where \(\sigma_{uv}\) is the covariance, and \(\sigma_{u}\) and \(\sigma_{v}\) are the standard deviations of the two quantities. The square (\(r_{uv}^2\)) represents what fraction of the variation in one quantity is explained by the other and gives a score for the validity of the hypothesis from \(0\) to \(1\). \(r_{uv}^2\) values close to \(1\) indicate a good hypothesis, between \(0.5\) and \(1\) a possibility, and below \(0.5\) very unlikely.
\subsection{Sample Selection}
\label{sec:orgf04047e}

Given the \ac{GP} model from the previous section, the algorithm searches within the region's bounds for the most informative location to sample. Sample locations are evaluated by an objective function based on Expected Improvement for Global Fit \cite{lam08sequen},
\begin{equation} \label{eq:objfunc}
\max_\mathbf{x} \hspace{2em} (\mu(\mathbf{x}) - y(\mathbf{x}^*))^2 + \alpha \: \sigma^2(\mathbf{x}) ,
\end{equation}
where \(\mathbf{x}\) is the chosen location, \(\mu(\mathbf{x})\) and \(\sigma^2(\mathbf{x})\) are respectively the \ac{GP} expected value and variance at the location, and \(y(\mathbf{x}^*)\) is the value of the nearest sample location. This function favors new samples that reduce a balance of local and global uncertainties. \(\alpha\) is a weight to choose this balance. A particle swarm optimization routine is used for the search.
\section{Results and Discussion}
\label{sec:org19d2563}

Synthetic data were used to focus on the methods being tested and to directly control relationships between quantities. Data for the \ac{QOI} were created by generating three to five Gaussian peaks located randomly within the search region. Secondary quantities were then generated from the \ac{QOI} using three methods:

\begin{itemize}
\item \textbf{High dependence (H)}: the entire map multiplied by a single value = \(QOI_i * \eta \; : \; \eta \sim \mathcal{N}(0, 1), \; \eta \ne 0\)
\item \textbf{Medium dependence (M)}: a different random value added at each point = \(QOI_i + 0.2 \eta_i \; : \; \eta_i \sim \mathcal{N}(0, 1)\)
\item \textbf{Low dependence (L)}: a completely new data map generated using the same randomized technique as the \ac{QOI}
\end{itemize}

\begin{figure}[t]
\centering
\includegraphics[width=1.0\linewidth]{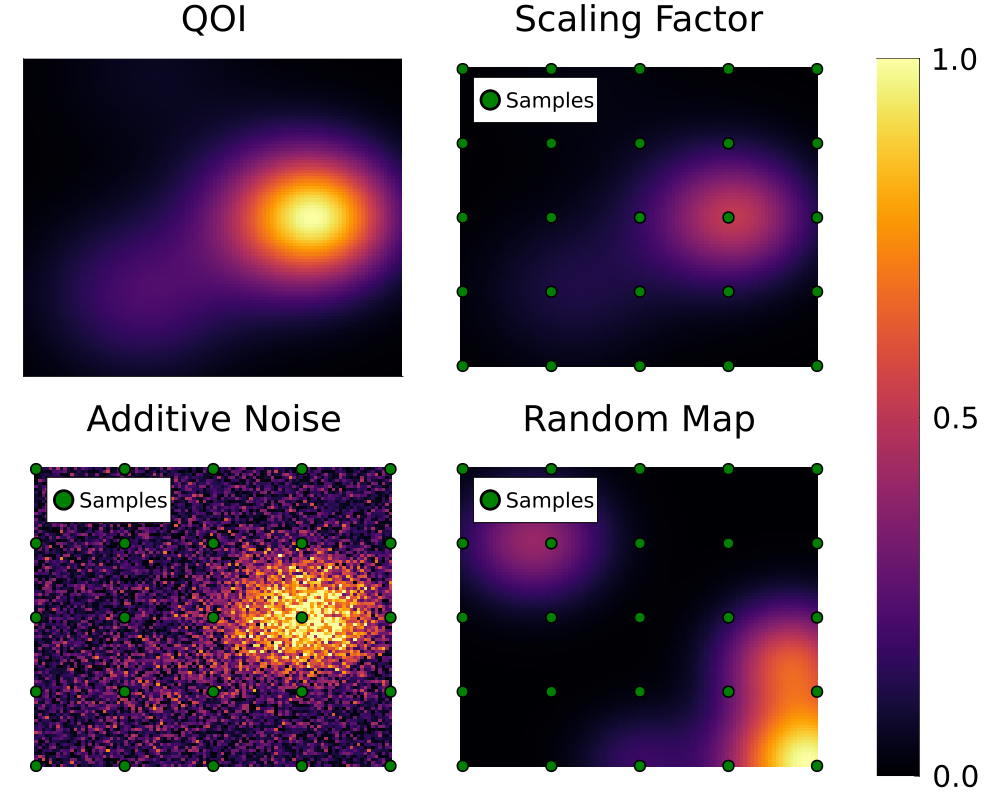}
\caption{\label{fig:env_data}Maps of the \ac{QOI} and prior quantities with varying levels of dependence. The top-right has high dependence with the \ac{QOI} (H), the bottom-left has medium dependence (M), and the bottom-right has low dependence (L).}
\end{figure}

To simulate having only sparse prior data, a 5x5 grid of samples evenly spread out over the region was extracted. See example maps in Fig. \ref{fig:env_data}.

The simulation was run as follows:
\begin{enumerate}
\item Initialize \ac{QOI} map, prior samples, search bounds, and starting location
\item Take sample at current location and train the model
\item Choose next sample location from the model
\item Continue taking samples and training until a budget of \(30\) samples is reached
\end{enumerate}

\begin{figure}[t]
\centering
\includegraphics[width=1.0\linewidth]{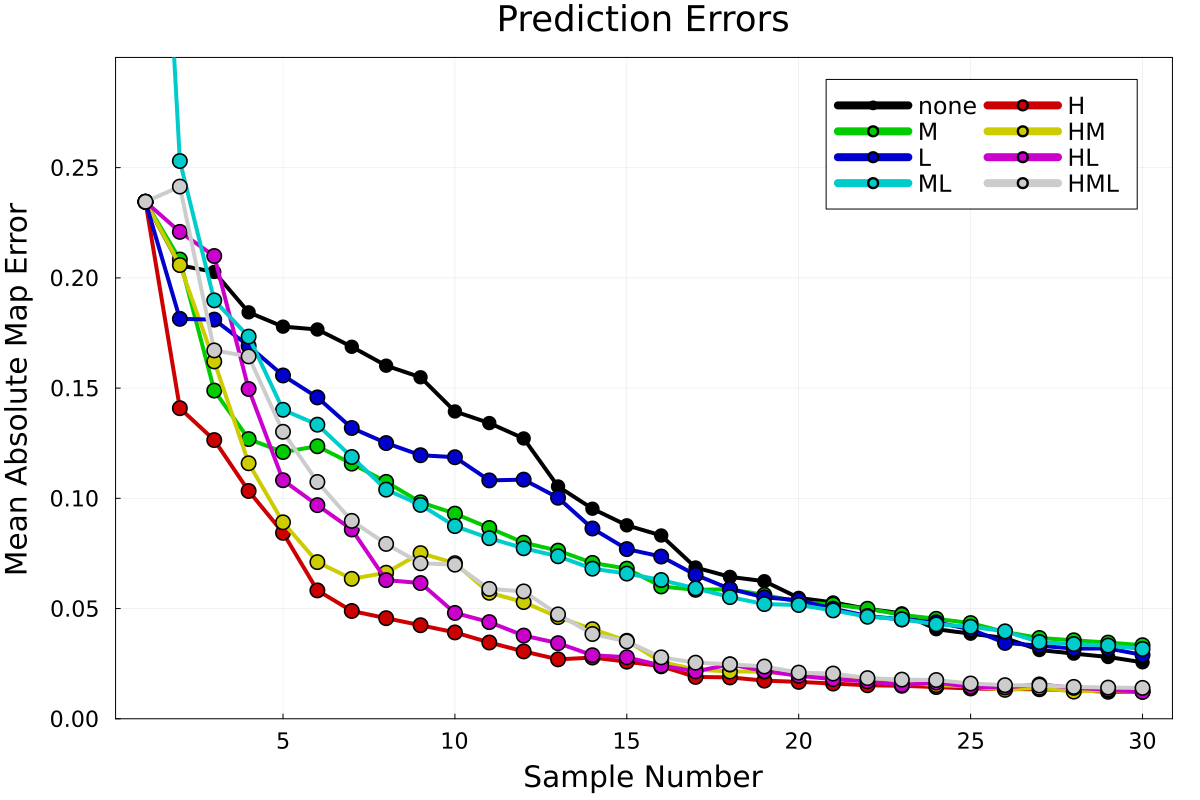}
\caption{\label{fig:errors}Mean prediction errors averaged over all runs. Each series is a different combination of High, Medium, and Low hypotheses. A division into three performance groups based on best available hypothesis is visible after \(7\) samples. The value not shown from the ML series is \(0.54\).}
\end{figure}

Simulations were run on all possible combinations of prior data from none to all three classes, making eight tests in all (e.g. see Fig. \ref{fig:errors}). This was repeated over \(18\) runs of randomized map data from which average performance was calculated. This was enough runs to get statistically significant differences between hypothesis groups (p < \(0.001\)). A value for \(\alpha\) of \(100\) was found to be a good balance with slightly more global exploration, as recommended in \cite{fuhg20stat_of}.

On a laptop AMD Ryzen 9 6900HS CPU, the simulation took \(0.26\) seconds on average to calculate both the belief model and next location after each sample. When run with all \(3\) priors with \(25\) samples each, it took \(2.1\) seconds on average. This can be reduced to below a second if less model refinement is chosen. These are acceptable times for in-situ prediction and decision-making and make real-time use possible.

\begin{figure}[t]
\centering
\includegraphics[width=1.0\linewidth]{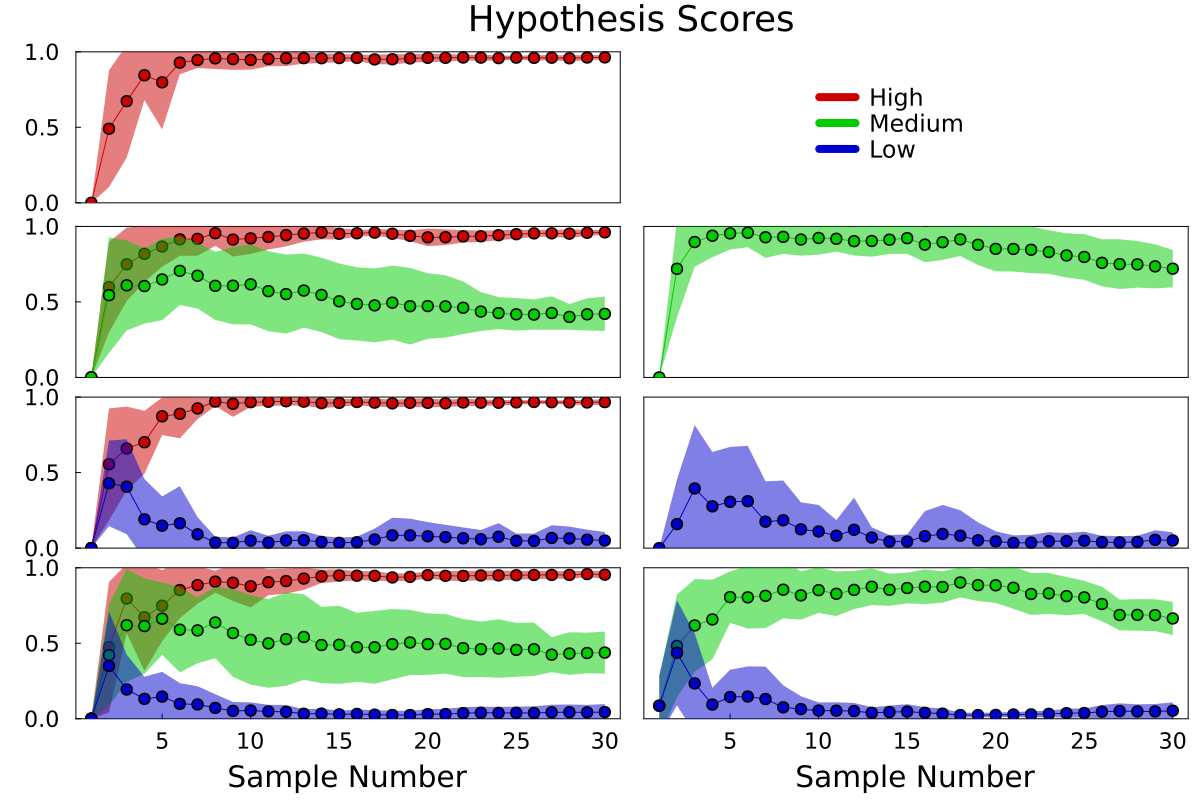}
\caption{\label{fig:scores}Mean hypothesis scores between each quantity and the \ac{QOI} throughout simulation runs of \(30\) samples. Each subplot is a different combination of High, Medium, and Low hypotheses with values averaged over all runs.}
\end{figure}

\subsection{Hypothesis Evaluations}
\label{sec:orge2cc8b0}

In Fig. \ref{fig:scores} we show the calculated hypothesis scores averaged over all simulation runs. For comparison, calculating corresponding scores for all true map values gave [\(1.0 \pm 0.0\), \(0.61 \pm 0.03\), \(0.06 \pm 0.06\)] for high, medium, and low across all the runs.

The algorithm had no problem in correctly evaluating the multiple hypotheses simultaneously. The three classes of prior data (H,M,L) all resulted in consistently distinct hypothesis scores from each other, no matter what the combination. The specific combinations did have slight effects on the scores, especially during the first \(5\) or so samples. For any method to robustly and accurately accept or reject hypotheses, sufficient data must be collected.

H-priors were assigned near-perfect scores after the first \(6\) samples and maintained this with high confidence throughout the rest of the run. This trend was consistent through all the possible combinations.

When not combined with a H-prior, M-priors started with higher scores that eventually came down to around \(0.83\). When H-priors were present, they more quickly settled to around \(0.45\). The existence of a good hypothesis helped to learn the distinction between it and the medium one.

L-priors universally scored below \(0.5\) after a few uncertain predictions. This class naturally had the greatest variance since the random map generation had more possibility to result in partial correlations.
\subsection{Accuracy to ground truth}
\label{sec:org009a617}

Fig. \ref{fig:errors} shows average prediction errors for all \(8\) prior data combinations. The algorithm adapted its use of the prior data, and by \(7\) samples the errors settled into three major groups. These groups are defined by the highest-dependence prior that they include.

Runs with a H-prior had the lowest errors and maintained that gap for the rest of the \(30\) samples. Inclusion of lower-dependence priors slightly degraded performance at the start, but by sample \(15\) they were all effectively equal.

The ML-priors run performed similarly to the M-prior run, and the L-prior run performed at least as well as when including no priors. Overall, each run performed no worse than its best hypothesis and incorrect hypotheses were quickly disregarded. Adding moderate to good hypotheses reduced prediction error and thus the number of samples needed to characterize the \ac{QOI}.

The principal benefits occur in the short to mid-term. Eventually, enough data of the \ac{QOI} will be gathered so that prior hypotheses no longer make a difference. This can be seen for runs with only a M-prior or worse: after around \(17\) samples, they all converged. It is expected that the same thing will occur for all the runs after many more samples. In other words, the \ac{QOI} will eventually be completely known, whether or not a high-quality hypothesis was used to start.

\begin{figure}[t]
\centering
\includegraphics[width=1.0\linewidth]{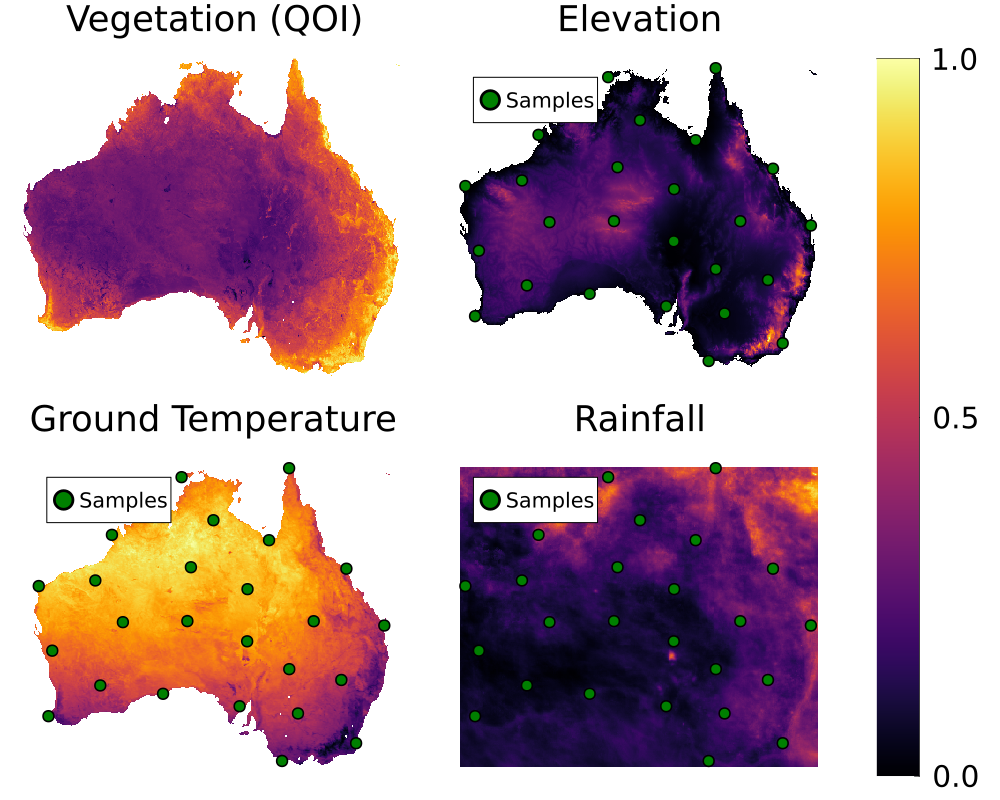}
\caption{\label{fig:aus_data_maps}Real data maps of vegetation, elevation, ground temperature, and rainfall throughout Australia and the sparse samples used. Values are averages over a year and have been normalized within each map.}
\end{figure}

\begin{figure}[t]
\centering
\includegraphics[width=1.0\linewidth]{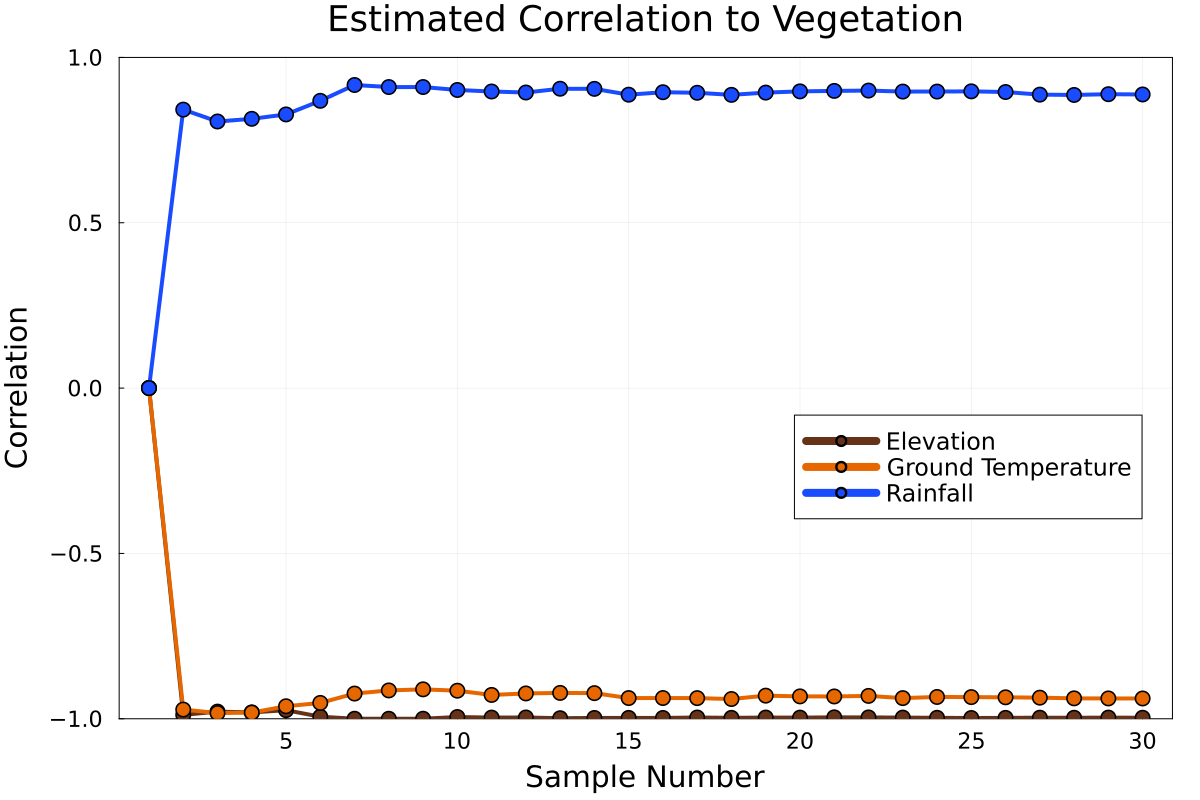}
\caption{\label{fig:aus_correlations}Estimated correlations to Vegetation over a run of \(30\) samples for Australia. It is estimated that rainfall is positively correlated while ground temperature and elevation are negatively.}
\end{figure}

\begin{figure}[t]
\centering
\includegraphics[width=1.0\linewidth]{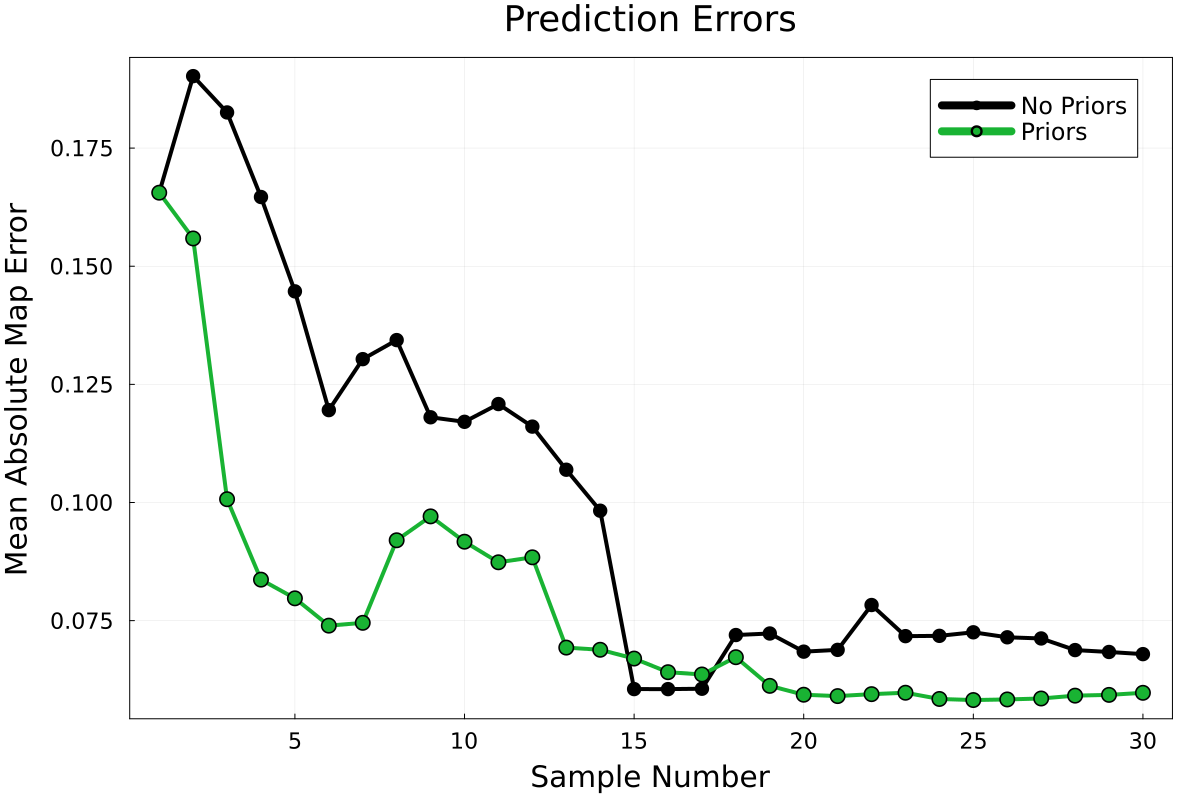}
\caption{\label{fig:aus_errors}Mean absolute prediction errors over a run of \(30\) samples for Australia. Using priors reduces errors over much of the run as compared to no priors.}
\end{figure}
\subsection{Real-World Examples}
\label{sec:orgf23eb30}

We further demonstrate the usefulness of these methods on two real-world datasets with real environmental relationships. This is to show the algorithmic methods of this paper working --- the scale and specific quantities can change per outdoor robotic application. Also note that the only thing simulated in these tests is the action of sampling --- the methods perform the same on the data regardless of how or when it was obtained. As a plus, having high-resolution offline data allows for ground truth comparisons not possible in the field.

\begin{figure}[t]
\centering
\includegraphics[width=1.0\linewidth]{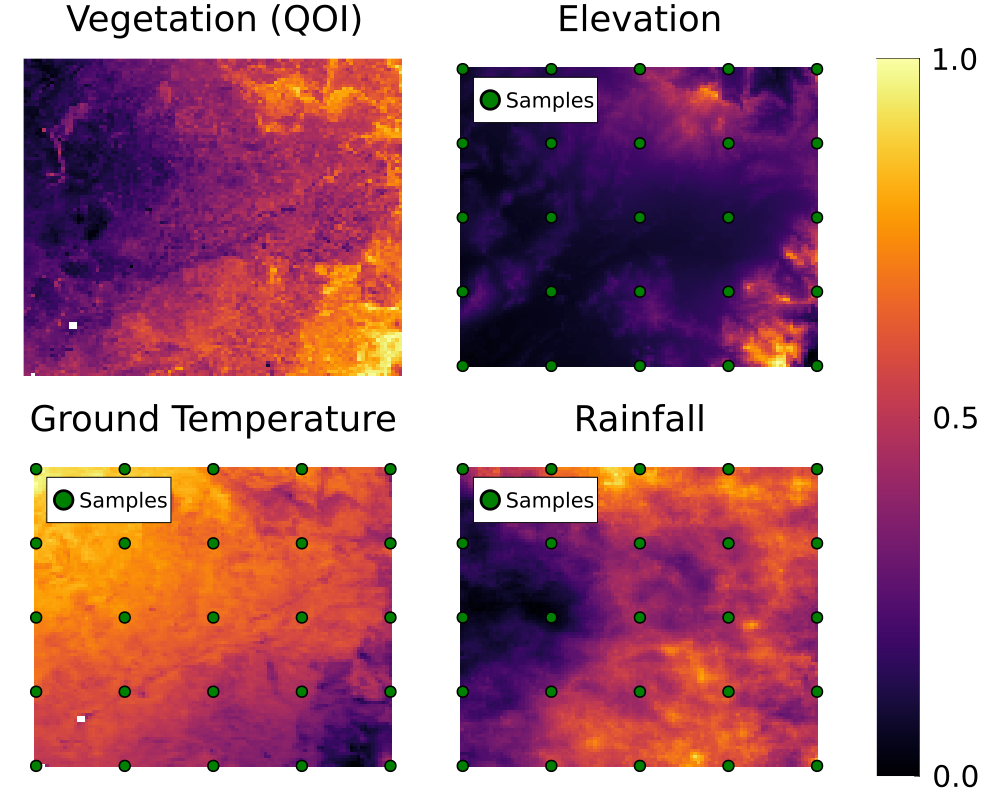}
\caption{\label{fig:nsw_data_maps}Real data maps of vegetation, elevation, ground temperature, and rainfall in a section of east Australia and the sparse samples used. Values are averages over a year and have been normalized within each map.}
\end{figure}

The data were taken from satellite imagery provided by NASA Earth Observations \cite{neo}. The two regions are the whole of Australia (Fig. \ref{fig:aus_data_maps}) and a 1000x1000 km patch of land between New South Wales and Queensland, Australia (Fig. \ref{fig:nsw_data_maps}). The \ac{QOI} was chosen to be vegetation and \(25\) dispersed samples were extracted from each of the other quantities for prior data. The algorithm was set to characterize vegetation across the map during a run of \(30\) samples.

After only \(2\) samples on the full Australia dataset, positive and negative correlations were identified and maintained for the rest of the run. Elevation and ground temperature were found to be highly negatively correlated and rainfall highly positively correlated (Fig. \ref{fig:aus_correlations}). Using these good hypotheses improved the prediction accuracy throughout almost the entire run, including up to \(75 \%\) better in the beginning (Fig. \ref{fig:aus_errors}).

Since the algorithm calculates a spatially-dependent correlation, a note on length-scale is important here. The amount of correlation in the data between two quantities changes depending on the length-scale of the trends examined. Our algorithm models this by the length-scale hyperparameter in the covariance function (\ref{eq:basefunc}), and its value is mainly determined by the distance between samples and the variation in their values. Our algorithm focuses on best global fit, so it favors more space between samples and hence a larger, more global length-scale. With that in mind, there may be local areas that differ from the global trend, especially with real data. For example, the negative correlation found between vegetation and elevation will not hold in the mountainous regions near the southeast coast. Development of strategies to capture both local and global trends, i.e. correlations on multiple length scales, is of interest for future work.

\begin{figure}[t]
\centering
\includegraphics[width=1.0\linewidth]{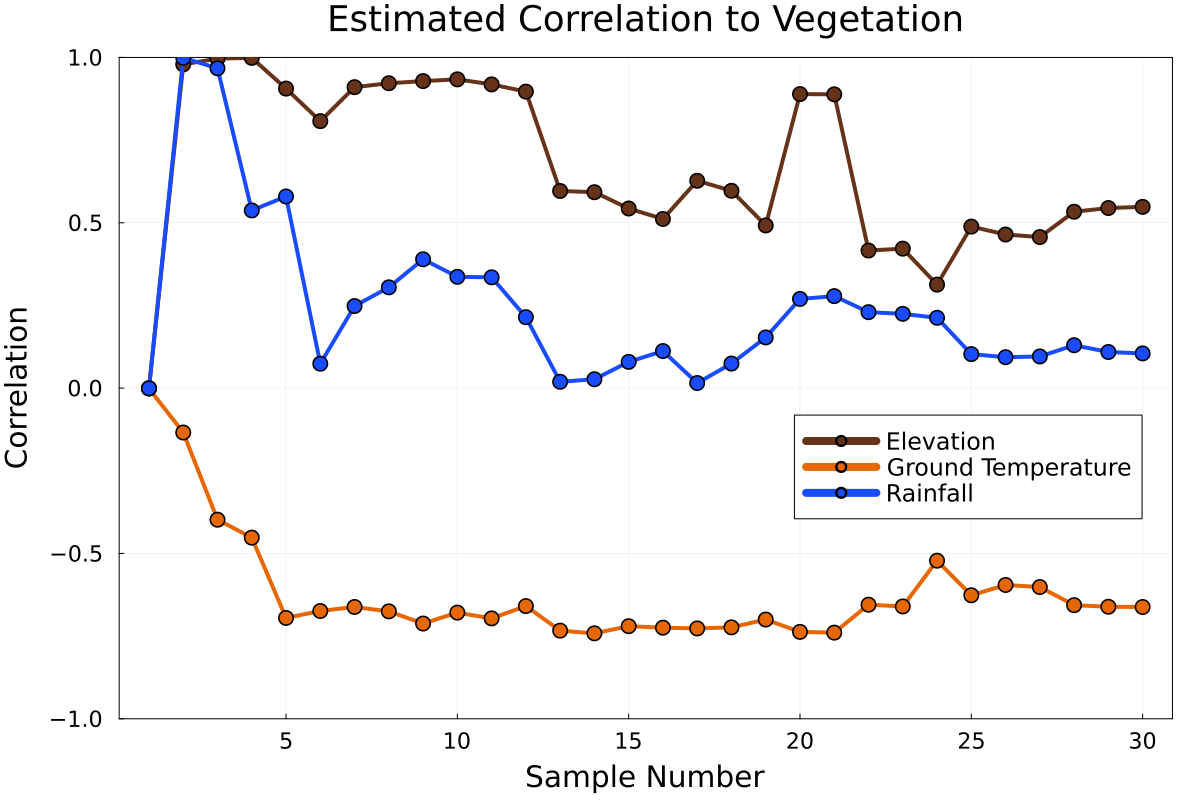}
\caption{\label{fig:nsw_correlations}Estimated correlations to Vegetation over a run of \(30\) samples. It is predicted that all quantities are partly correlated with vegetation.}
\end{figure}

\begin{figure}[t]
\centering
\includegraphics[width=1.0\linewidth]{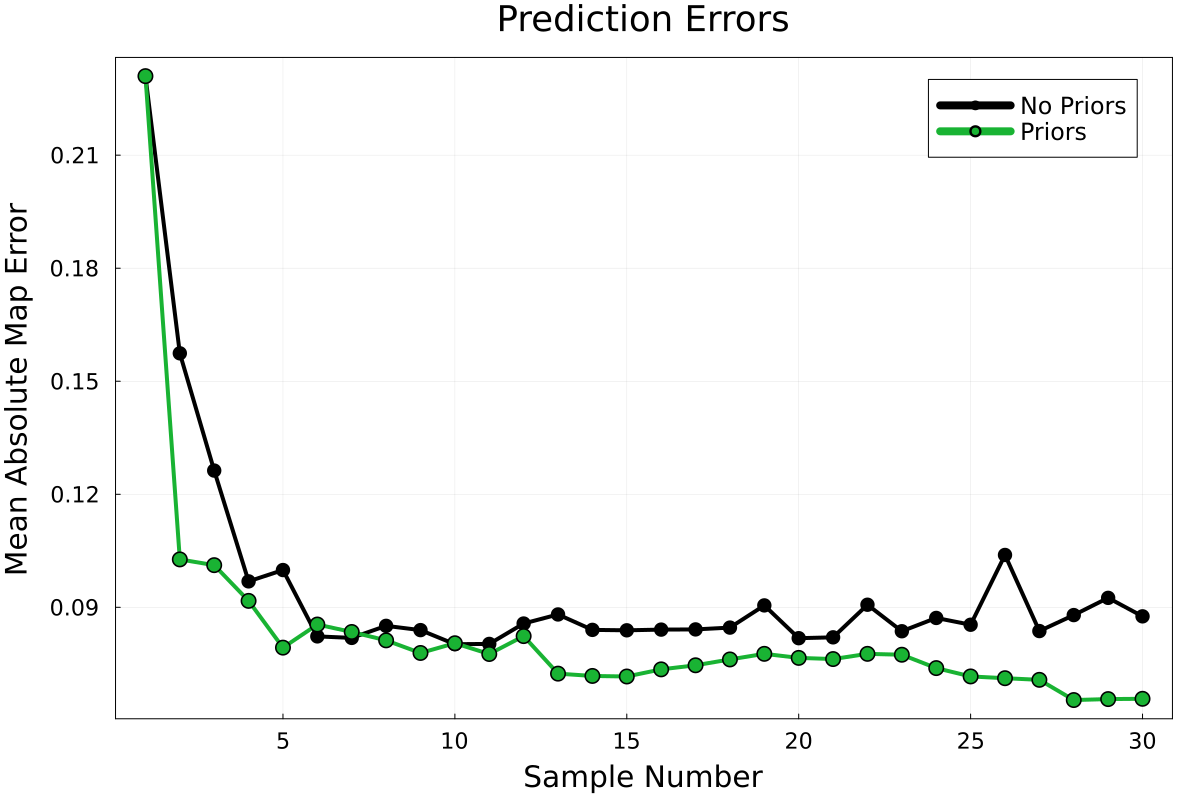}
\caption{\label{fig:nsw_errors}Mean absolute prediction errors over a run of \(30\) samples. Using priors slightly improves over using no priors, particularly at the end.}
\end{figure}

For the local dataset, the method established correlation estimates by around \(7\) samples, and they remained mostly stable over the course of the run. The results were partial positive correlations with elevation and rainfall and a negative correlation with ground temperature (Fig. \ref{fig:nsw_correlations}). Mean prediction errors showed slightly improved performance when using priors compared with using no priors (Fig. \ref{fig:nsw_errors}).

This used a subset of the overall map from the first example, and it demonstrates what was mentioned in the discussion: local correlations may be different than global ones. At this smaller-scale view, the effect of the mountains on the surrounding climate is more apparent in the data and higher elevations are found to be correlated with more vegetation. It's also important to note again that this is a spatially-dependent correlation rather than a point-by-point one. This means similarities are allowed to be found in regions around each point based on the learned length-scale. As with any model, the underlying assumptions must be understood and assessed as to their applicability for the given scenario. Deeper explanations still fall to the scientist, but the key here is that the algorithm aids them in their analysis.

\section{Conclusion}
\label{sec:orgb48e8d7}

This paper presented a method to address two related tasks in outdoor applications: evaluating spatially-dependent environmental hypotheses and reducing the number of samples to characterize a \ac{QOI}. The method simultaneously evaluates the quality of multiple hypotheses about linear dependence between quantities after every sample. Furthermore, it adapts its learning to the quality of each hypothesis: good hypotheses increase the amount of information gained from each sample and poor hypotheses lose effect after a few samples. Hypotheses identified as high-quality can more confidently be used in further outdoor applications.

This system is suitable for in-situ information gathering on robotic platforms, though the use of this method with real data requires extra considerations. Effective operation is currently limited to linear correlations on a global scale. Future work will investigate nonlinear correlations and correlations on multiple length-scales.

\bibliographystyle{IEEEtran}
\bibliography{references}

\begin{thebibliography}{10}
\providecommand{\url}[1]{#1}
\csname url@rmstyle\endcsname
\providecommand{\newblock}{\relax}
\providecommand{\bibinfo}[2]{#2}
\providecommand\BIBentrySTDinterwordspacing{\spaceskip=0pt\relax}
\providecommand\BIBentryALTinterwordstretchfactor{4}
\providecommand\BIBentryALTinterwordspacing{\spaceskip=\fontdimen2\font plus
\BIBentryALTinterwordstretchfactor\fontdimen3\font minus
  \fontdimen4\font\relax}
\providecommand\BIBforeignlanguage[2]{{%
\expandafter\ifx\csname l@#1\endcsname\relax
\typeout{** WARNING: IEEEtran.bst: No hyphenation pattern has been}%
\typeout{** loaded for the language `#1'. Using the pattern for}%
\typeout{** the default language instead.}%
\else
\language=\csname l@#1\endcsname
\fi
#2}}

\bibitem{kwon19effec_nitr}
S.-J. Kwon, \emph{et~al.}, ``Effects of nitrogen, phosphorus and potassium
  fertilizers on growth characteristics of two species of bellflower
  (platycodon grandiflorum),'' \emph{Journal of Crop Science and
  Biotechnology}, vol.~22, pp. 481--487, 2019.

\bibitem{rasm06gaus_proc}
C.~E. Rasmussen and C.~K.~I. Williams, \emph{Gaussian Processes for Machine
  Learning}.\hskip 1em plus 0.5em minus 0.4em\relax Massachusetts Institute of
  Technology, 2006.

\bibitem{gunt12effec_mois}
\BIBentryALTinterwordspacing
M.~Gunti{\~n}as, \emph{et~al.}, ``Effects of moisture and temperature on net
  soil nitrogen mineralization: a laboratory study,'' \emph{European Journal of
  Soil Biology}, vol.~48, pp. 73--80, 2012. [Online]. Available:
  \url{https://www.sciencedirect.com/science/article/pii/S1164556311000732}
\BIBentrySTDinterwordspacing

\bibitem{zhuan21comp_surv}
F.~Zhuang, \emph{et~al.}, ``A comprehensive survey on transfer learning,''
  \emph{Proceedings of the IEEE}, vol. 109, no.~1, pp. 43--76, 2021.

\bibitem{Weiss_2016}
\BIBentryALTinterwordspacing
K.~Weiss, \emph{et~al.}, ``A survey of transfer learning,'' \emph{Journal of
  Big Data}, vol.~3, no.~1, May 2016. [Online]. Available:
  \url{http://dx.doi.org/10.1186/s40537-016-0043-6}
\BIBentrySTDinterwordspacing

\bibitem{furl18fora_pros}
P.~M. Furlong, ``Foraging, prospecting, and falsification-improving three
  aspects of autonomous science,'' 2018.

\bibitem{burg20mobil_robot}
B.~Burger, \emph{et~al.}, ``A mobile robotic chemist,'' \emph{Nature}, vol.
  583, no. 7815, pp. 237--241, Jul 2020.

\bibitem{aror18onlin_mult}
A.~Arora, \emph{et~al.}, ``Online multi-modal learning and adaptive informative
  trajectory planning for autonomous exploration,'' in \emph{Field and Service
  Robotics}, M.~Hutter and R.~Siegwart, Eds.\hskip 1em plus 0.5em minus
  0.4em\relax Cham: Springer International Publishing, 2018, pp. 239--254.

\bibitem{marc12bayes_intel}
R.~Marchant and F.~Ramos, ``Bayesian optimisation for intelligent environmental
  monitoring,'' in \emph{2012 IEEE/RSJ International Conference on Intelligent
  Robots and Systems}, 10 2012, p. nil.

\bibitem{edel20ergod_traj}
K.~Edelson, ``Ergodic trajectory optimization for information gathering,''
  Master's thesis, Carnegie Mellon University, Pittsburgh, PA, October 2020.

\bibitem{cand20plan_rover}
A.~Candela, \emph{et~al.}, ``Planetary rover exploration combining remote and
  in situ measurements for active spectroscopic mapping,'' in \emph{2020 IEEE
  Intl. Conf. on Robotics and Automation (ICRA)}, 2020, pp. 5986--5993.

\bibitem{fuhg20stat_of}
J.~N. Fuhg, \emph{et~al.}, ``State-of-the-art and comparative review of
  adaptive sampling methods for kriging,'' \emph{Archives of Computational
  Methods in Engineering}, vol.~28, no.~4, pp. 2689--2747, Aug 2020.

\bibitem{shah16takin_human}
\BIBentryALTinterwordspacing
B.~Shahriari, \emph{et~al.}, ``Taking the human out of the loop: a review of
  bayesian optimization,'' \emph{Proceedings of the IEEE}, vol. 104, no.~1, pp.
  148--175, 2016. [Online]. Available:
  \url{http://dx.doi.org/10.1109/JPROC.2015.2494218}
\BIBentrySTDinterwordspacing

\bibitem{leen90empl_elev}
H.~Leenaers, \emph{et~al.}, ``Employing elevation data for efficient mapping of
  soil pollution on floodplains,'' \emph{Soil Use and Management}, vol.~6,
  no.~3, pp. 105--114, 1990.

\bibitem{meht21adap_samp}
M.~Mehta and C.~Shao, ``Adaptive sampling design for multi-task learning of
  gaussian processes in manufacturing,'' \emph{Journal of Manufacturing
  Systems}, vol.~61, no. nil, pp. 326--337, 2021.

\bibitem{boni08mult_task}
E.~V. Bonilla, \emph{et~al.}, ``Multi-task gaussian process prediction,'' in
  \emph{Advances in Neural Information Processing Systems}, J.~Platt,
  \emph{et~al.}, Eds., vol.~20.\hskip 1em plus 0.5em minus 0.4em\relax Curran
  Associates, Inc., 2008.

\bibitem{lam08sequen}
C.~Q. Lam, ``Sequential adaptive designs in computer experiments for response
  surface model fit,'' Ph.D. dissertation, The Ohio State University, 2008.

\bibitem{neo}
\BIBentryALTinterwordspacing
NEO, ``Dataset index.'' [Online]. Available:
  \url{https://neo.gsfc.nasa.gov/dataset_index.php}
\BIBentrySTDinterwordspacing

\end{thebibliography}
\end{document}